# Assistive robotic device: evaluation of intelligent algorithms

Audrey Lebrasseur[1,2], Josiane Lettre[2], François Routhier[1,2], Philippe Archambault[3,4], Alexandre Campeau-Lecours[1,2]

[1]Université Laval, Quebec City, Canada, [2]Centre for interdisciplinary research in rehabilitation and social integration, CIUSSS de la Capitale-National, Quebec City, Canada, [3]Centre for interdisciplinary research in rehabilitation of Greater Montreal, Centre intégré de santé et de services sociaux de Laval, Laval, Canada, [4]McGill University, Montreal, Canada.

## ABSTRACT

Assistive robotic devices can be used to help people with upper body disabilities gaining more autonomy in their daily life. Although basic motions such as positioning and orienting an assistive robot gripper in space allow performance of many tasks, it might be time consuming and tedious to perform more complex tasks. To overcome these difficulties, improvements can be implemented at different levels, such as mechanical design, control interfaces and intelligent control algorithms. In order to guide the design of solutions, it is important to assess the impact and potential of different innovations. This paper thus presents the evaluation of three intelligent algorithms aiming to improve the performance of the JACO robotic arm (Kinova Robotics). The evaluated algorithms are 'preset position', 'fluidity filter' and 'drinking mode'. The algorithm evaluation was performed with 14 motorized wheelchair's users and showed a statistically significant improvement of the robot's performance.

## INTRODUCTION

Robots for assisting in manipulation constitute a powerful solution for enhancement of autonomy for people living with severe upper-limb neuromotor impairments. Such robots may assist people in various tasks, such as eating, drinking, moving various objects, turning book pages, opening and closing doors and turning on and off electrical and electronic systems. They also serve to move the user's hand or feet, to cross legs or to stretch and can be used as a rehabilitation device in clinics (Van der Loos & Reinkensmeyer, 2008). Lately, a new generation of multitask assistive robots was introduced. Some of these robotic manipulators were designed to be installed on a fixed workstation while others were onboard mobile devices that followed users in their daily activities. The latter offered more flexibility to the users and allowed them to enhance their autonomy for a wider spectrum of activities. While many onboard robotic manipulators were developed, very few were commercialized. Some examples of assistive robots are the iARM by Exact Dynamics which resulted from the MANUS project (Kwee, 2000) and the Raptor by Phybotics. These assistive robots should allow efficient operations with most activities of daily living (ADL), comply with the environmental limits of a motorized wheelchair, possess an intuitive control, and allow further development and integration of new tools.

JACO is a robotic arm produced by Kinova Robotics (www.kinovarobotics.com) that consists of a fixed base (controller) which is linked to 6 carbon fiber shells and a gripper through rotating actuators. It is designed to be installed on a motorized wheelchair (Figure 1) and used by people living with upper extremity mobility limitations. Controlled through the wheelchair drive control, it allows the user to reach, move and manipulate objects in his/her surroundings. This allows enhanced autonomy in daily activities such as drinking, scratching an itch, eating, picking up items, opening doors and more (Maheu et al., 2011). Different clinical trials have been conducted in the last few years and have shown that this assistive device is easy to use in several daily tasks considered as important for the users (Routhier & Archambault, 2010), increases the users' autonomy (Clark, 2013) and social participation (Routhier et al., 2014), and reduces the required caregiving time (Maheu et al., 2011).

JACO's Cartesian controller allows 7 basic movements: 3 translations of the gripper in space, 3 rotations of the wrist and opening/closing of the fingers. A 'mode' button selectors allows users to switch between controlling the gripper's position, the wrist rotation or the opening and closing of fingers. In addition to these movements, the Home function sends JACO, through a single command, to a position where it is ready to use or to rest. Although basic motions such as positioning and orienting the gripper in space allow performance of many tasks, the users' needs are not quite met. Indeed, complex or repetitive tasks may take too long to achieve and may also be very tedious. This could be due to multiple factors such as few available independent control signals, lack of fine movements to perform precision tasks such as picking up a straw and severe user's spasms. To overcome these difficulties, different solutions were proposed in the literature, such as alternative control interfaces (accelerometers (Fall et al., 2015), EMG (Côté-Allard et al., 2017), computer-vision (Ka et al., 2016)) and control algorithms (Vu et al., 2017; Herlant et al., 2016; Campeau-Lecours et al., 2016).

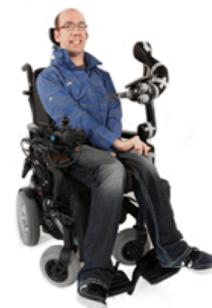

**Figure 1: JACO mounted on a power wheelchair.**

## OBJECTIVES

This paper presents the evaluation of three algorithms aiming to improve the usability and the performance of the robotic assistive device JACO. The evaluated algorithms are 'preset position', 'fluidity filter' and 'drinking mode' (see below and also Campeau-Lecours (2016) for a detailed description).

## METHODS

### Participants and recruitment

This study was approved by the Research Ethics Board at the *Centre intégré universitaire de santé et de services sociaux de la Capitale-Nationale* (CIUSSS-CN) and an informed consent was obtained from each participant. Participants were recruited through the mobility aids program of the CIUSSS-CN and a list of people who participated in other projects of the co-authors and accepted to be contacted for future projects. In order to fulfill this research project's objective, the algorithms were tested by people with upper limbs disabilities, either with a non-functional shoulder's range of motion or with a non-functional grasping. The participants had to be between 18 and 65 years old and had to be powered wheelchair users since at least six months.

### Variables

Three main components were evaluated for each algorithm, namely: 1) the usability (with the A Questionnaire for the Evaluation of Physical Assistive Devices (QUEAD) which evaluates the subjective performance of physical assistive devices (Schmidtler, 2017)), 2) the easiness of accomplishment (on a 7-point Likert scale, "1" being "Very difficult" and "7" being "Very easy"), and 3) the task completion time (except for the drinking mode algorithm, as detailed in the results' section).

### Procedure

Every participant was involved in an individual session of approximately two and a half hours at the *Centre for interdisciplinary research in rehabilitation and social integration, Quebec City, Canada*. The JACO robotic arm was installed on a table placed in front of the participant. The device needed to be on the opposite side of the powered wheelchair's joystick to respect its usual installation conditions. The control board, which contained the joystick and the four buttons required to manipulate the robotic arm, was either placed on the participant' lap (when feasible) or on the table. Five tasks were designed to test the three control algorithms (see below). All tasks were accomplished with and without assistance from the control algorithms; half of the participants started the tasks with the algorithms and the other half started without. Moreover, the order with which the algorithms were evaluated was randomized to prevent any bias (due, for instance, to the participant's energy level or to practice).

### Algorithm A: Preset Position (Tasks 1 & 2)

Many simple tasks, such as pushing on an elevator button, or elaborate tasks, such as eating, are time consuming and complex if performed with an assistive device like JACO. Those tasks necessitate the user to toggle through multiple modes to access many different motion commands. The idea of the preset position was to enable the user to record a position or multiple positions (trajectory). Once the position defined by the participant, the user can reach them with a simple command (e.g. pushing a button) and possibly optimize the time and effort needed to perform the task. Two tasks were defined to evaluate this feature. Task 1 consisted in reaching a button located 60 cm in front, 30 cm to the left and 100 cm upward from the starting position (to simulate pushing an elevator button). Task 2 consisted in beginning from a starting position, reaching above a plate on a table, and reaching to a position near the mouth (to simulate eating with the robot). These tasks were performed with and without the algorithm. In the latter case, users were asked to reach the position by using a joystick with basic functionalities (left, right, up, etc.).

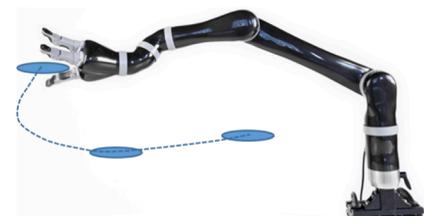

Figure 2: JACO preset position algorithm.

### Algorithm B: Fluidity Filter (Tasks 3 & 4)

Users living with important muscular spasms might accidently induce errors in control signals while moving the joystick. Filtering the control signal is a simple method to reduce the impact of these involuntary commands.

However, while traditional filters help to decrease the acceleration, they also filter the deceleration, which can lead to a counter intuitive behavior. Indeed, with traditional filters, when the user has reached the target, the device continues to move and overshoots the target. A fluidity filter was thus designed to limit the acceleration of JACO without limiting the deceleration (Campeau-Lecours, 2016). Two tasks (#3 & 4) were defined to evaluate this feature. Task 3 consisted in grabbing a glass of water on a table and Task 4 was to grab a pen from a pencil holder. Tasks were performed with the fluidity filter, with a first order low pass filter and without filter.

### Algorithm C: Drinking Mode (Task 5)

Most of JACO users drink with a straw. However, for those who wish to drink directly from a glass or to pour a drink, a combination of upward and backward translations along with a rotation is required at the gripper location. While able-bodied individuals easily perform this motion, it is a complex motion to perform with a robot. To overcome this barrier, an automatic synchronization of the drinking motion was designed. For the study, the participants were asked to drink water directly from a glass with and without the algorithm.

### Analysis

A descriptive quantitative analysis was first used to evaluate the data. Mean results of the time taken, the ease of use and the answers of the questionnaire were calculated and analyzed with Excel (Version 15.13.1). Moreover, we performed an analysis of variance (ANOVA) to determine the influence of the algorithms (with vs without). The p values were calculated with nparLD from R Studio (Version 2.1). The level of significance was set at 0.05.

## RESULTS

Table 1 presents the demographic characteristics of the 14 participants. Table 2 presents the results for the preset position. Task 1 refers to the task of reaching a button while Task 2 refers to reaching a plate and mouth back and forth. Table 3 presents the results of the fluidity filter test. Task 3 refers to grabbing a glass of water while Task 4 refers to grabbing a pencil. Table 4 presents the results for the drinking mode. In this case, because many participants were not able to accomplish the task at all without the algorithm, we did not evaluate the task completion time since it was obviously not a meaningful measure.

**Table 1 – Demographic characteristics of the participants**

| Characteristic | n |
|---|---|
| Sex | |
| Male | 8 |
| Female | 6 |
| Age | |
| 18-34 | 4 |
| 35-54 | 5 |
| 55 + | 5 |
| Diagnosis | |
| Tetraplegia | 6 |
| Spinal muscular atrophy | 4 |
| Other | 4 |

**Table 2 – Preset Position Evaluation Results**

| | Task | With algo | Without algo | Diff (%) | p |
|---|---|---|---|---|---|
| Task Completion time (sec.) | 1 | 24.4 | 75.0 | -67.4 | 0.0002* |
| | 2 | 35.9 | 129.9 | -72.4 | 0.0001* |
| Easiness (1 to 7) | 1 | 6.6 | 5.5 | 20.8 | 0.01* |
| | 2 | 6.8 | 4.5 | 50.8 | 0.002* |
| Usability (1 to 7) | 1 and 2 | 6.2 | 5.1 | 21.0 | 0.0002* |

* Statistically significant

**Table 3 – Fluidity Filter Evaluation Results**

| | Task | With algo | Classic | Without | p |
|---|---|---|---|---|---|
| Task Completion time (seconds) | 3 | 76.0 | 89.1 | 109.7 | 0.2 |
| | 4 | 132.7 | 139.2 | 137.0 | 0.5 |
| Easiness (1 to 7) | 3 | 6.4 | 5.5 | 5.5 | 0.1 |
| | 4 | 5.2 | 4.9 | 5.2 | 0.9 |
| Usability (1 to 7) | 3 and 4 | 5.9 | 4.5 | 5.7 | 0.1 |

## DISCUSSION

In this paper, we have presented the evaluation of different algorithms that were developed to increase JACO's performances. The 'preset position' algorithm led to a significant reduction of task completion time of 67.4% and 72.4% for task 1 and 2, respectively. This was accompanied by a significant increase in the perceived easiness of control between 20.8% and 50.8% and to a significant increase in usability perception of 21% on average for both tasks. The 'fluidity filter' algorithm also led to a reduction of task completion time and increased appreciation. However, the results were not statistically significant. The small sample size can explain this result. While different JACO's users mentioned this algorithm as very useful, no user in our sample had neuromuscular impairments such as spasticity or tremors, which may have reduced the importance of such algorithms. Finally, the 'drinking mode' algorithm showed a statistically significant increase of 51.9% of the perceived easiness to complete the task and an increase of 25.5% for the usability perception.

**Table 4 – Drinking Mode Evaluation Results**

| | Task | With algo | Without algo | Diff (%) | p |
|---|---|---|---|---|---|
| Easiness (1 to 7) | 5 | 5.9 | 3.9 | 51.9 | 0.01* |
| Usability (1 to 7) | 5 | 5.9 | 4.7 | 25.5 | 0.002* |

## CONCLUSION

Although JACO's basic functionalities allow the user to perform many tasks, advanced functionalities were required to further empower the users. This paper has presented the evaluation of three algorithms that were specifically designed to increase JACO users' capacities as well as the number of achievable tasks, while decreasing the time and effort needed to achieve them. In general, the evaluation has shown that the proposed algorithms might have an important statistically significant impact on the task completion time, the easiness of accomplishment and on the usability perception of the JACO robotic arm. The evaluation of the algorithms leads to the conclusion that the development of intelligent control algorithms is an interesting path to improve the performance of assistive robotic devices.

## ACKOWLEDGEMENTS

This project was funded by Pr. Campeau-Lecours start-up funds from CIRRIS.

## REFERENCES

Campeau-Lecours, A., Maheu, V., Lepage, S., Lamontagne, H., Latour, S., Paquet, L., Hardie, N. (2016). Jaco assistive robotic device: Empowering people with disabilities through innovative algorithms. Rehabilitation Engineering and Assistive Technology Society of North America (RESNA).

Côté-Allard, U., Fall, C.L., Campeau-Lecours, A., Gosselin, C., Laviolette, F., Gosselin, B. (2017). Transfer Learning for sEMG Hand Gesture Recognition Using Convolutional Neural Networks. IEEE International Conference on Systems, Man, and Cybernetics.

Clark E. Utilisation d'un bras robotisé par des personnes ayant un contrôle moteur diminué aux membres supérieurs. (2013) Motricité Cérébrale : Réadaptation, Neurologie du Développement. 34: 63-70.

De Santis, A., Siciliano, B., De Luca, A., Bicchi, A. (2008). An atlas of physical human–robot interaction. Mechanism and Machine Theory. 43(3): 253-270.

Fall, C. L., Turgeon, P., Campeau-Lecours, … Gosselin, B. (2015). Intuitive Wireless Control of a Robotic Arm for people living with an upper body Disability. IEEE Intern. Conf. Engineering in Medicine and Biology Society.

Herlant, L. V., Holladay, R. M., Srinivasa, S. S. (2016). Assistive teleoperation of robot arms via automatic time-optimal mode switching. International Conference on Human-Robot Interaction (HRI).

Ka, H., Ding, D., Cooper, R. A. (2016). Three Dimentional Computer Vision-Based Alternative Control Method for Assistive Robotic Manipulator. Symbiosis, 1(1)

Kwee H.H. (2000). Integrated control of MANUS manipulator and wheelchair enhanced by environmental docking. Robotica 16: 491-498.

Maheu, V., Frappier, J., Archambault, PS., Routhier, F. (2011). Evaluation of the JACO robotic arm: Clinico-economic study for powered wheelchair users with upper-extremity disabilities. IEEE International Conference on Rehabilitation Robotics. 1-5.

Routhier, F., Archambault, PS. (2010). Usability of a joystick-controlled six degree-of-freedom robotic manipulator. Rehabilitation Engineering and Assistive Technology Society of North America, RESNA Annual Conference.

Routhier, F., Archambault, P., Cyr, M. C., Maheu, V., Lemay, M., Gélinas, I. (2014). Benefits Of Jaco Robotic Arm On Independent Living And Social Participation: An Exploratory Study. Rehabilitation Engineering and Assistive Technology Society of North America, RESNA Annual Conference.

Schmidtler, J., Bengler, K., Dimeas, F., Campeau-Lecours, A. (2017). A Questionnaire for the Evaluation of Physical Assistive Devices (QUEAD). IEEE International Conference on Systems, Man, and Cybernetics.

Van der Loos, HF. M., Reinkensmeyer, DJ. (2008). Rehabilitation and health care robotics. Springer Handbook of Robotics. 1223-1251.

Vu, D. S., Allard, U. C., Gosselin, C., Routhier, F., …, Campeau-Lecours, A. (2017). Intuitive adaptive orientation control of assistive robots for people living with upper limb disabilities. Intern. Conf. on Rehabilitation Robotics.